\documentclass{article}
\usepackage{spconf,amsmath,epsfig}
\usepackage{booktabs}
\usepackage{graphicx}
\usepackage{subcaption}
\usepackage{amssymb} 
\title{SAR to optical image translation with color supervised diffusion model}
\name{Xinyu Bai\textsuperscript{1,2}, Feng Xu\textsuperscript{2}}

\address{\textsuperscript{1}Yiwu Research Institute of Fudan University,\\
        \textsuperscript{2}Key Lab for Information Science of Electromagnetic Waves (MoE),\\
        Fudan University, Shanghai 200433, China, \\
        fengxu@fudan.edu.cn}
\begin{document}
%\ninept
%
\maketitle
\begin{abstract}
Synthetic Aperture Radar (SAR) offers all-weather, high-resolution imaging capabilities, but its complex imaging mechanism often poses challenges for interpretation. In response to these limitations, this paper introduces an innovative generative model designed to transform SAR images into more intelligible optical images, thereby enhancing the interpretability of SAR images. Specifically, our model backbone is based on the recent diffusion models, which have
powerful generative capabilities. We employ SAR images as conditional guides in the sampling process and integrate color supervision to counteract color shift issues effectively. We conducted experiments on the SEN12 dataset and employed quantitative evaluations using peak signal-to-noise ratio, structural similarity, and fréchet inception distance. The results demonstrate that our model not only surpasses previous methods in quantitative assessments but also significantly enhances the visual quality of the generated images.
\end{abstract}
\section{Introduction}
\label{sec:intro}
Synthetic Aperture Radar (SAR) represents a cutting-edge technology in the field of remote sensing, employing a combination of radar principles and synthetic aperture techniques to facilitate high-resolution imaging. SAR images, notable for their high resolution and ground penetration capabilities, serve as a continuous and reliable source of terrestrial information. The evolution of SAR in recent years has been marked by significant advancements, including the development of multi-polarization, multi-baseline, multi-frequency, multi-angle, and multi-channel systems. These advancements have substantially enhanced SAR image quality, making SAR an indispensable tool in diverse fields such as climate change research and environmental monitoring.

Despite the advantages of all-weather, high-resolution imaging of SAR, interpreting its images is often complicated by speckle noise and geometric distortions\cite{fu2021reciprocal}. Conversely, optical images offer clearer spectral and spatial information but struggle under cloud cover and atmospheric scattering. Bridging this gap, converting SAR to optical images enhances SAR's interpretability and leverages the strengths of both imaging modalities. While traditional methods depend on costly machine learning and complex feature extraction, recent advances in deep learning, particularly Generative Adversarial Networks (GANs)\cite{goodfellow2020generative} and Denoising Diffusion Probabilistic Models (DDPMs)\cite{ho2020denoising}, have revolutionized this conversion process.

Recently, diffusion models have emerged as a mainstream generative method, surpassing current GAN-based generation models in image synthesis\cite{dhariwal2021diffusion}. In the generation process of DDPMs, clean images are produced by starting with random Gaussian noise and iteratively denoising. Diffusion models are trained by maximizing the variational lower bound of the negative log-likelihood\cite{ho2020denoising}, mitigating the occurrence of mode collapse commonly in GANs. However, due to the lack of color supervision during training, the generated images often exhibit color shift issues\cite{10330015}. Therefore, this paper introduces a color-supervised conditional diffusion model with a $1 \times 1$ convolution structure for extracting information from SAR images. This approach, compared to direct connections, more effectively utilizes the information in SAR images.

In summary, the contributions of this paper are twofold:
(1) We introduce an innovative conditional diffusion model featuring color supervision. Our approach effectively preserves the target information, making the boundaries clearer and greatly reducing the overall color shift.
(2) We conduct experiments on the SEN12 dataset to visually demonstrate the transformation capabilities. Furthermore, we compared our model against other existing models, establishing its superiority in both quantitative and qualitative assessments. 
\begin{figure}[!t]
  \centering
   \includegraphics[width=0.8\linewidth]{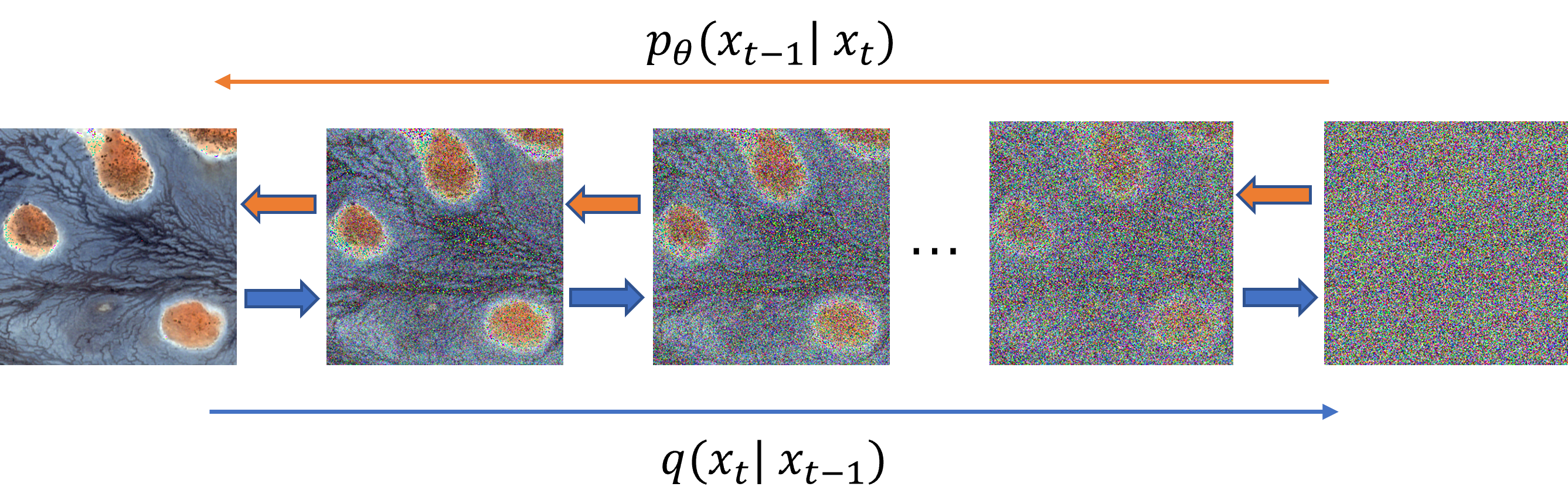}
   
   \caption{Overview of the diffusion process.}
   \label{fig:pr}
   \vspace{-1em}
\end{figure}

\section{Related Work}
\label{sec:rw}
\vspace{-1em}
\subsection{SAR-to-Optical Image Translation}
\vspace{-0.5em}
The interpretation of SAR images is a critical yet challenging task in remote sensing, with a growing interest in converting SAR to more interpretable optical images. Traditional methods have focused on colorizing SAR images to differentiate objects but often fall short in depicting the actual ground scenario. Fu et al.\cite{fu2021reciprocal} refined this approach with a multi-scale discriminator for better image synthesis. Despite the progress, GANs often struggle with training difficulties and blurry boundaries. Bai et al.\cite{10330015} attempted to overcome these with DDPM, yet faced issues with color accuracy. Addressing these shortcomings, we introduce a conditional diffusion model with color loss supervision to enhance the fidelity and clarity of SAR to optical image translation.
\vspace{-1em}
\subsection{Diffusion Models}
\vspace{-0.5em}
Recently developed diffusion models have emerged as advanced generative tools, surpassing GANs in various computer vision tasks. Dhariwal et al.\cite{dhariwal2021diffusion} proposed some modifications to the original DDPMs, such as learning the variances of the reverse process, using a different noise schedule, and applying a residual connection to the denoising function.  Offering greater diversity, training stability, and scalability than GANs, diffusion models represent a significant advancement in generative technology. This paper harnesses the diffusion model, using SAR images as a guiding condition, to adeptly transform SAR into high-quality optical images, showcasing the potent image generation capabilities of diffusion models.

\begin{figure}[!t]
  \centering
   \includegraphics[width=\linewidth]{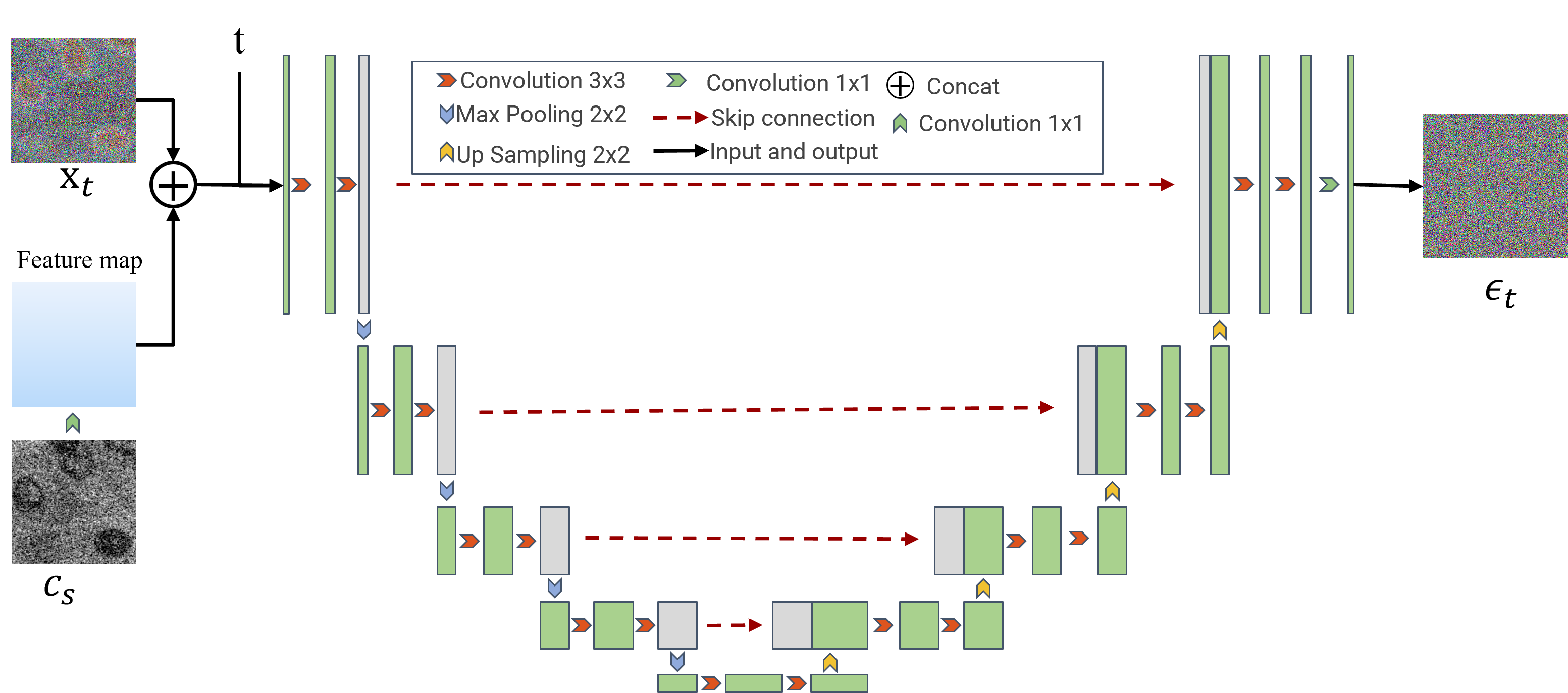}
   \caption{Optical image generation conditioned on SAR images. The SAR image $c_{s}$ is dimensionally increased by a $1 \times 1$ convolution.}
   \label{fig:full}
\end{figure}

\section{Methods}
\label{sec:mt}
\vspace{-0.5em}
We design a SAR-to-optical translation model based on the diffusion model with color supervision.
Section \ref{sec:pl} briefly introduces the original diffusion model, and Section \ref{sec:mtd} introduces our proposed methods.

\subsection{Preliminaries}
\label{sec:pl}
\vspace{-0.5em}
Our diffusion model architecture is built upon the framework delineated by \cite{ho2020denoising}, which is composed of forward and backward processes. Both of these are Markov chains, as depicted in Fig.\ref{fig:pr}.
During the forward diffusion phase, the original image $x_{0}$ undergoes transformation into $x_{T} \sim N(0,1)$ through the progressive infusion of Gaussian noise $\epsilon$ across $T$ diffusion steps, according to a predefined variance schedule $\left \{ \beta _{1},\beta _{2}, \cdots, \beta _{T} \right \} $:
\begin{equation}
q\left(x_{t} \mid x_{t-1}\right):=N\left(x_{t} ; \sqrt{1-\beta_{t}} x_{t-1}, \beta_{t} \mathbf{I}\right)
\end{equation}
By defining $\alpha_{t}=1-\beta_{t}$ and $ \bar{\alpha}_{t}=\prod_{t=1}^{T} \alpha_{i}$, it becomes feasible to sample $x_{t}$ at any chosen time step $t$. Specifically, $x_t$ can be obtained by the following formula:
\begin{equation}\label{func:3}
x_t = \sqrt{\bar{\alpha}_{t}} x_{0}+\sqrt{1-\bar{\alpha}_{t}} \epsilon
\end{equation}
As direct modeling of the reverse diffusion is intricate, the model is designed to parameterize the Gaussian transformation $p_{ \theta }\left (x_{t-1} \mid x_{t} \right )$. Essentially, the model forecasts the mean of the Gaussian distribution $\mu_{\theta}(x_{t}, t)$, while the variance $\sigma_t$ remains constant.
Therefore, the backward diffusion process is delineated as:
\begin{equation}
p_\theta(x_{t-1}|x_t)=N(x_{t-1};\mu_\theta(x_t,t),\sigma_t^2\mathbf{I})
\end{equation}
Ho et al.\cite{ho2020denoising} have derived a simplified version of the loss function:
\begin{equation}
L_{simple}=\mathbb{E}_{t,x_{0},\epsilon}\big[\|\epsilon-\epsilon_\theta(\mathbf{x}_t,t)\|^2\big]
\end{equation}
where $\epsilon_\theta(\mathbf{x}_t,t)$ denotes the Gaussian noise as parameterized by the mean of the prediction.

\subsection{SAR-Guided Diffusion with Color Supervision}
\label{sec:mtd}
We propose a novel approach for converting SAR images into their optical counterparts, utilizing a conditional diffusion model enhanced with color supervision. As depicted in Fig.\ref{fig:full}, our method employs SAR images as the condition to steer the generation process, underpinned by a U-net architecture tasked with noise prediction.

Our core idea is to leverages SAR images to guide the sampling process of the model. Specifically, in the forward diffusion process, clean optical images $x_0$ serve as the starting point. These images undergo a transformation into $x_t$ through the addition of Gaussian noise. Concurrently, the SAR image $c_s$ is upsampled using a $1 \times 1$ convolution, resulting in a three-channel feature map $fm$.
The model's main backbone receives a concatenated input of $x_t$ and $fm$. The reverse generation process initiates from Gaussian noise 
$x_T$ and gradually denoises, all the while being conditioned on the SAR image, to progressively morph into the corresponding optical image. Importantly, the SAR image conditions, applied at each diffusion step, are noise-free and have been processed through a $1 \times 1$ convolution structure, ensuring refined and consistent guidance throughout the model's operation. The modified conditional generation is represented as:
\begin{equation}
p_\theta(x_{t-1}|x_t,c_{s})=N(x_{t-1};\mu_\theta(x_t,t,c_{s}),\sigma_t^2\mathbf{I})
\end{equation}
Regarding the loss function, we incorporate an extra color loss component. Building upon Eq.\ref{func:3}, we deduce the expression for $x_0$ when we get the $\epsilon$:
\begin{equation}
x'_0 = \tfrac{1}{\sqrt{\bar\alpha }}(x_t-\sqrt{1-\bar\alpha}\epsilon)
\end{equation}
The $x'_0$ subsequently undergoes Gaussian blurring\cite{ignatov2017dslr}. This blurring technique selectively obscures boundary details while preserving the general color profile. This processed component is then integrated with the inherent DDPM loss to constitute our comprehensive final loss function:
\begin{equation}
L = \mathbb{E}_{t,x_{0},\epsilon}\big[\|\epsilon-\epsilon_\theta(x_t,t,c_{s})\|^2\big] + \mathbb{E}_{x_{0}}\big[\|x_{0b}-x'_{0b}\|^2\big]
\end{equation}

\begin{figure*}[ttt]
\centering
    \includegraphics[width=0.13\linewidth]{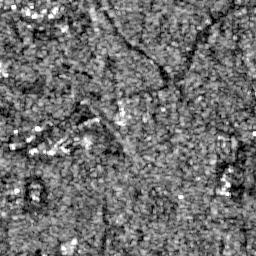}
  \hfil
   \includegraphics[width=0.13\linewidth]{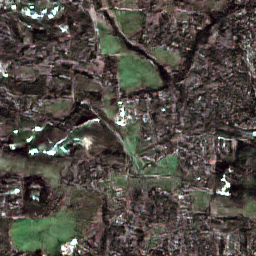}
  \hfil
   \includegraphics[width=0.13\linewidth]{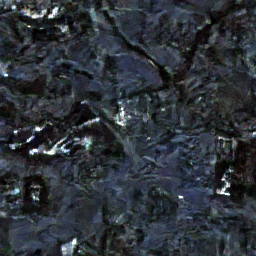}
  \hfil
    \includegraphics[width=0.13\linewidth]{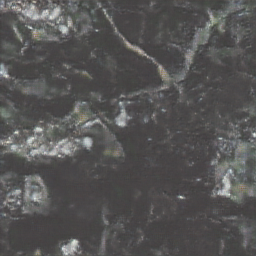}
  \hfil
    \includegraphics[width=0.13\linewidth]{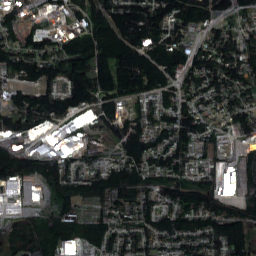}
  \hfil
    \includegraphics[width=0.13\linewidth]{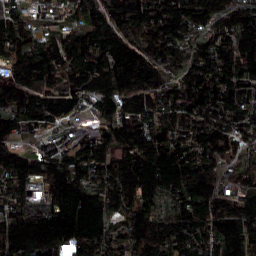}
  \hfil
    \includegraphics[width=0.13\linewidth]{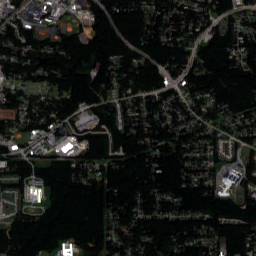}
  \hfil
\\[0.1cm]
    \includegraphics[width=0.13\linewidth]{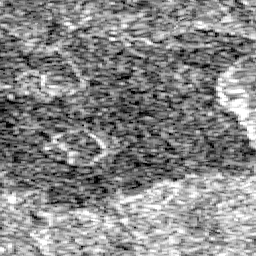}
  \hfil
   \includegraphics[width=0.13\linewidth]{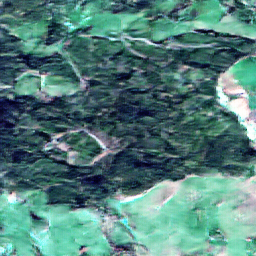}
  \hfil
   \includegraphics[width=0.13\linewidth]{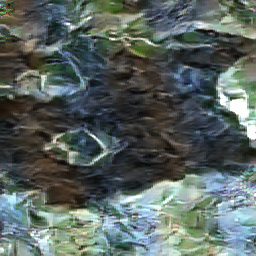}
  \hfil
    \includegraphics[width=0.13\linewidth]{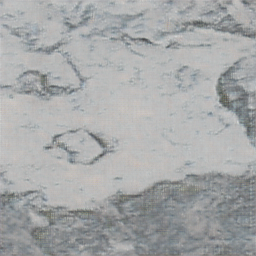}
      \hfil
    \includegraphics[width=0.13\linewidth]{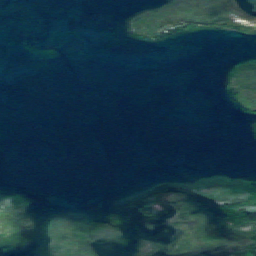}
  \hfil
    \includegraphics[width=0.13\linewidth]{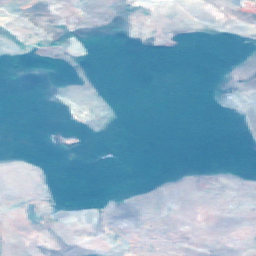}
  \hfil
    \includegraphics[width=0.13\linewidth]{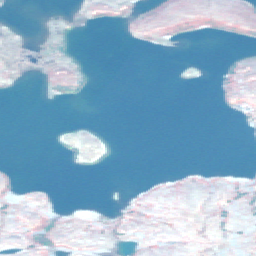}
  \hfil
\\[0.1cm]
    \includegraphics[width=0.13\linewidth]{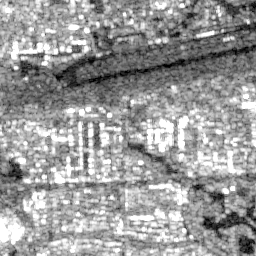}
  \hfil
   \includegraphics[width=0.13\linewidth]{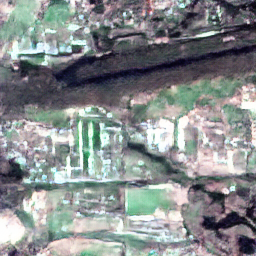}
  \hfil
   \includegraphics[width=0.13\linewidth]{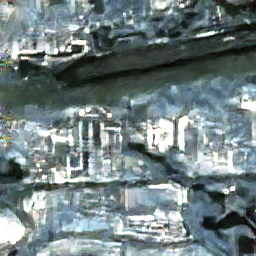}
  \hfil
    \includegraphics[width=0.13\linewidth]{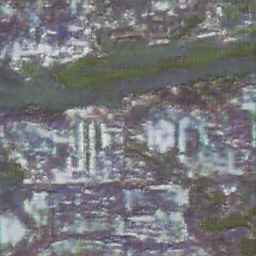}
      \hfil
    \includegraphics[width=0.13\linewidth]{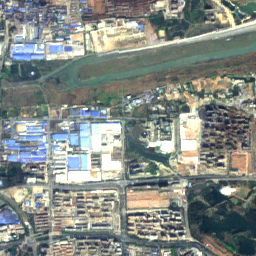}
  \hfil
    \includegraphics[width=0.13\linewidth]{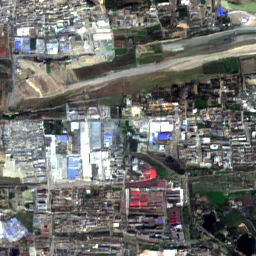}
  \hfil
    \includegraphics[width=0.13\linewidth]{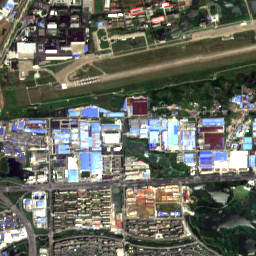}
\\[0.1cm]
\hspace{-0.16cm}
   \subfloat[SAR\strut]{ \includegraphics[width=0.13\linewidth]{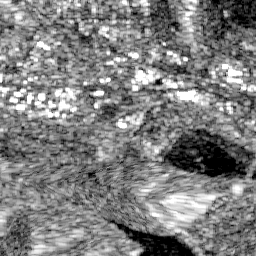}}
  \hspace{0.14cm}
  % \hfil
   \subfloat[CycleGAN\strut]{\includegraphics[width=0.13\linewidth]{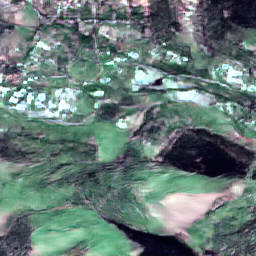}}
  \hspace{0.14cm}
  % \hfil
   \subfloat[NiceGAN\strut]{\includegraphics[width=0.13\linewidth]{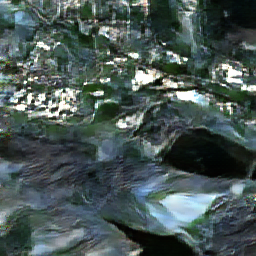}}
  \hspace{0.14cm}
  % \hfil
    \subfloat[CRAN\strut]{\includegraphics[width=0.13\linewidth]{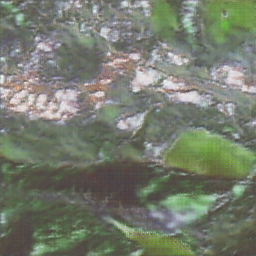}}
  \hspace{0.05cm}
  % \hfil
    \subfloat[S2ODPM\strut]{\includegraphics[width=0.13\linewidth]{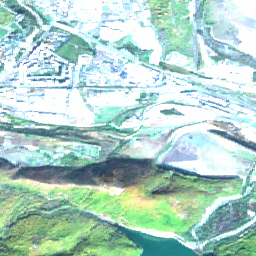}}
  \hspace{0.12cm}
  % \hfil
   \subfloat[Ours\strut]{ \includegraphics[width=0.13\linewidth]{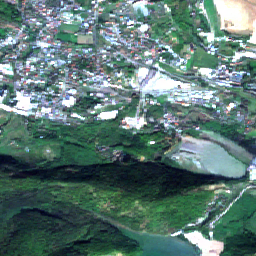}}
  \hspace{0.12cm}
  % \hfil
   \subfloat[GT\strut]{\includegraphics[width=0.13\linewidth]{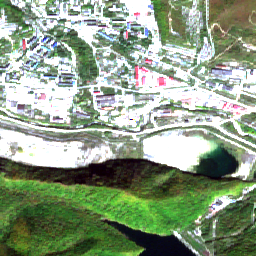}}
   \vspace{-0.5em}
\caption{Translation results across various models. Each column corresponds to: (a) the SAR image; (b) the result generated by CycleGAN, (c) NiceGAN, (d) CRAN, (e) S2ODPM, (f) our proposed model, and (g) the real optical image (ground truth).}
\label{fig:cp2}
\end{figure*}
\vspace{-0.5em}
\section{Experiment}
\label{sec:exp}

\subsection{Dataset and Metrics}
We conduct experiments on the satellite SEN12 dataset\cite{schmitt2018sen1} to demonstrate its performance. 
The SEN12 dataset consists of 282,384 SAR and optical image pairs acquired by Sentinel-1 and Sentinel-2, respectively, which are collected from locations across the landmasses and seasons.

For the quantitative evaluation, we use Peak signal-to-noise ratio (PSNR), Structural Similarity (SSIM) and Fréchet Inception Distance (FID)\cite{heusel2017gans} to measure the quality of the generated optical images. 
\vspace{-0.7em}
\subsection{Implementation Details}
\vspace{-0.5em}
In this work, we set $T = 1000$.
We train the model with the AdamW optimizer with $\theta_1=0.9$ and $\theta_2=0.999$. The learning rate is linearly increased from 0 to 5e-5. We run the model for 80,000 iterations on 4 × 24G NVIDIA 3090 GPUs with a batch size of 24.
\vspace{-0.5em}
\begin{table}[htbp]
\centering
\caption{Result comparisons of different methods.}
\label{tab:mytable}
\begin{tabular}{cccc}
\toprule
Method & PSNR$\uparrow$ & FID$\downarrow$ & SSIM$\uparrow$ \\
\midrule
CycleGAN & 13.33 & 192.13 & 0.2156  \\
NiceGAN & 12.48 & 211.62 & 0.2181  \\
CRAN & 14.13 & 214.46 & 0.2289  \\
S2ODPM & 18.76 & 136.38 & 0.3020  \\
\textbf{Ours} & \textbf{19.72} & \textbf{116.93} & \textbf{0.3119} \\
\bottomrule
\end{tabular}
\end{table}

\subsection{Experiment Results}
In our comprehensive evaluation of the proposed method against established benchmarks in image translation, we conducted a comparative analysis with several models to underscore the efficacy of our approach. Specifically, CycleGAN\cite{zhu2017unpaired} and NiceGAN\cite{chen2020reusing} were chosen for their previously advanced performance in image-to-image translation tasks. Additionally, we incorporated CRAN\cite{fu2021reciprocal}, a novel adversarial network optimized for SAR to optical image conversion, utilizing cascading residual connections and a hybrid L1-GAN loss. Furthermore, we included S2ODPM\cite{10330015}, a model predicated on DDPM principles, which has recently achieved excellent performance in SAR to optical image translation.

To ensure fairness, we employed the official implementations of these models and conducted experiments on the SEN12 dataset under identical conditions. The results, as presented in Table I, show that our method outperformed the competition across key metrics such as PSNR, SSIM, and FID. These metrics collectively indicate a superior capacity of our conditional diffusion model to accurately learn and replicate the complex mapping rules intrinsic to SAR and optical image conversion. 

For a more concrete illustration of our model's capabilities, we undertook an extensive visual analysis. We selected a diverse set of scenes from the SEN12 dataset and generated their optical image counterparts using our model. These results were then visually compared with those generated by other models. As demonstrated in Figure 3, our model's outputs are noticeably more realistic and natural-looking. The delineation of different objects is starkly clearer, and the overall color fidelity is markedly improved. In comparison to GAN-based models, our DDPM-based approach yields images that are not only more accurate in detail but also exhibit fewer artifacts, offering a cleaner, more coherent optical image. When pitted against the S2ODPM model, our approach exhibited substantial improvements in overall color reproduction and target differentiation. These improvements are pivotal, as they greatly enhance the model's ability to facilitate precise and reliable interpretations in practical applications. However, a notable drawback of our model is its slower inference speed due to the need to iteratively infer $T$ times to generate images. 
\vspace{-0.5em}
\section{Conclusion}
\vspace{-1em}
In this paper, we introduce an innovative SAR to optical image translation model that leverages a diffusion process augmented with color supervision. The approach transforms Gaussian noise into optical images, guided by SAR images. Our model stands out for producing optical images with enhanced clarity and significantly reduced color shift, surpassing the existing models. Moreover, it adeptly circumvents the prevalent issue of mode collapse typically encountered during the training phase. Quantitative evaluations validate the model's superior performance, particularly noting its ability to preserve structural consistency between the original SAR and the synthesized optical images. Future work will focus on how to accelerate the model's sampling speed.

% Below is an example of how to insert images. Delete the ``\vspace'' line,
% uncomment the preceding line ``\centerline...'' and replace ``imageX.ps''
% with a suitable PostScript file name.
% ------------------------------------------------------------------------

% To start a new column (but not a new page) and help balance the last-page
% column length use \vfill\pagebreak.
% -------------------------------------------------------------------------
\vfill
\pagebreak

% References should be produced using the bibtex program from suitable
% BiBTeX files (here: strings, refs, manuals). The IEEEbib.bst bibliography
% style file from IEEE produces unsorted bibliography list.
% -------------------------------------------------------------------------
\bibliographystyle{IEEEbib}
\bibliography{main}

\end{document}